\documentclass[11pt]{article}
\usepackage[margin=1in]{geometry}

\usepackage{graphicx}
\usepackage{booktabs}

\usepackage{amsmath,amssymb}
\usepackage{multirow}
\usepackage{array}
\usepackage{enumitem}
\usepackage{authblk}
\usepackage{xcolor}
\usepackage{hyperref}
\usepackage{siunitx}

\graphicspath{{figures/}}

\begin{document}

\title{Retrieval-Augmented Online Learning for Concept Drift Adaptation}
\author[1]{Wenzhang Du\thanks{Correspondence: \texttt{dqswordman@gmail.com}}}
\affil[1]{Department of Computer Engineering, International College (MUTIC), Mahanakorn University of Technology, Bangkok, Thailand}
\date{}
\maketitle

\begin{abstract}
Online learning in non-stationary environments is difficult because data distributions drift and previously trained models degrade. Classical approaches either forget old data with sliding windows or fading factors, or rely on drift detectors that trigger model resets. Forgetting discards recurring patterns, while detectors are fragile and often react late.

We address this setting with retrieval-augmented memory. At each time step we maintain a small buffer of past examples, retrieve those whose representations are closest to the current input, and jointly train on the current example and its neighbours. This provides an additional, data-dependent gradient signal that can accelerate adaptation when new regimes resemble previously seen ones.

Building on this idea, we introduce \textbf{RAM-OL} (Retrieval-Augmented Memory for Online Learning), a plug-in module for standard online learners. We study a naïve replay variant and a gated variant that down-weights stale or dissimilar memories via a time window, similarity gating and gradient re-weighting. On three public data streams with concept drift---ElecNormNew, multi-site electricity load and airline delay prediction---RAM-OL matches or improves prequential accuracy over a purely online multilayer perceptron baseline and substantially reduces variance across seeds. A stylised analysis under bounded drift clarifies when retrieval can reduce adaptation cost compared with purely online learning.
\end{abstract}

\section{Introduction}

Machine learning models deployed in dynamic environments rarely satisfy the i.i.d.\ assumption. User preferences evolve, sensors drift and markets change, so the data distribution varies over time. This \emph{concept drift} leads to gradual but sometimes severe performance degradation unless models adapt~\cite{WidmerKubat1996,GamaEtAl2014}. Online learning updates parameters after each new observation, but fast and robust adaptation under drift remains challenging.

Most existing approaches fall into two families~\cite{GamaEtAl2014}. Passive adaptation methods use sliding windows or exponential fading to down-weight old data, or dynamic ensembles whose weights evolve over time~\cite{Klinkenberg2004,KolterMaloof2007}. They track smooth drift but discard historical patterns, so recurring regimes must be re-learned when they reappear. Active drift detectors, such as ADWIN, monitor error statistics and trigger resets when statistically significant change is detected~\cite{BifetGavalda2007}. These methods are sensitive to thresholds, prone to false alarms or missed detections, and often react after accuracy has already dropped.

We explore a complementary perspective inspired by human long-term memory: instead of only forgetting or resetting, an online learner should selectively \emph{remember} and reuse relevant past experience. At each step we maintain a small buffer of past examples, retrieve those whose representations are most similar to the current input and jointly update on the current example and its neighbours. This retrieval-augmented mechanism is related to experience replay in reinforcement learning~\cite{Lin1992,MnihEtAl2015}, exemplar-based continual learning~\cite{RebuffiEtAl2017,RiemerEtAl2019} and retrieval-augmented models in NLP~\cite{LewisEtAl2020,GuuEtAl2020,BorgeaudEtAl2022}, but here it is tailored to streaming supervised learning under concept drift with an explicit focus on safety.

We propose Retrieval-Augmented Memory for Online Learning (\textbf{RAM-OL}), realised as a lightweight extension of online stochastic gradient descent (SGD). Naïve replay can help but may also harm adaptation if it overuses outdated neighbours. We therefore introduce a gated replay variant that constrains neighbours via a time window, similarity gating and gradient re-weighting. We instantiate RAM-OL on top of a simple online multilayer perceptron (MLP) and evaluate it on three standard data streams derived from the Elec2 electricity pricing benchmark~\cite{Harries1999} and public repositories~\cite{BacheLichman2013,BTSAirline}. RAM-OL improves over a purely online baseline on two challenging electricity datasets and matches it on a noisy airline stream.

Our contributions are:
\begin{itemize}[leftmargin=*]
    \item We propose \textbf{RAM-OL}, a retrieval-augmented online learning framework that can be added to standard gradient-based learners without architectural changes, together with design principles for safe memory usage under concept drift.
    \item We provide a theoretical perspective via a bounded-drift model and sketch regret upper and lower bounds~\cite{Zinkevich2003,HazanSeshadhri2007,BesbesEtAl2015} that clarify when retrieval can reduce adaptation cost relative to purely online learning.
    \item We present a systematic empirical study on three real data streams, showing that RAM-OL increases accuracy and markedly reduces seed variance on two electricity datasets while never underperforming the baseline on a noisy airline stream. Ablation studies disentangle the contributions of different gating components.
\end{itemize}

\section{Related Work}
\label{sec:related}

\subsection{Online learning and concept drift}

Concept drift has long been recognised as a central challenge for inductive learning in dynamic environments~\cite{WidmerKubat1996}. Passive adaptation schemes restrict effective training data to a recent horizon using sliding windows, decaying weights or dynamic ensembles~\cite{Klinkenberg2004,KolterMaloof2007}. Drift detectors such as ADWIN~\cite{BifetGavalda2007} adjust window sizes based on statistical tests. A comprehensive survey is given by Gama et al.~\cite{GamaEtAl2014}. These approaches work well for smooth, one-way drift but do not explicitly exploit reoccurring regimes: once an old concept has been forgotten, it must be re-learned when it returns.

Non-stationary online learning has also been analysed from a regret perspective. Zinkevich~\cite{Zinkevich2003} established foundational results for online convex programming. Subsequent work introduced variation budgets and adaptive regret, yielding bounds that scale as $O(\sqrt{T} + V_T)$ where $V_T$ measures the total amount of change~\cite{HazanSeshadhri2007,BesbesEtAl2015}. These analyses typically assume that examples are observed once and then discarded; explicit memory and retrieval mechanisms are rarely considered. Our discussion in Section~\ref{sec:theory} builds on this line of work.

\subsection{Experience replay and retrieval-augmented models}

Experience replay is a standard tool in reinforcement learning, where agents store and re-sample transitions to stabilise training~\cite{Lin1992,MnihEtAl2015}. In class-incremental and continual learning, stored exemplars are used to mitigate catastrophic forgetting~\cite{RebuffiEtAl2017,RiemerEtAl2019}. Most replay strategies sample buffers uniformly or according to simple heuristics.

Retrieval-augmented models in NLP combine parametric predictors with non-parametric memories queried via nearest-neighbour search. Representative examples include retrieval-augmented generation~\cite{LewisEtAl2020}, REALM~\cite{GuuEtAl2020} and retrieval-based large language models~\cite{BorgeaudEtAl2022}. These systems retrieve from large text corpora, often at inference time.

RAM-OL borrows the nearest-neighbour retrieval idea but applies it in a streaming supervised setting, with a small, local buffer and an explicit emphasis on avoiding harm under unfavourable drift.

\section{Problem Setup}
\label{sec:setup}

We consider online supervised learning with concept drift. At each discrete time step $t = 1, 2, \dots, T$:
\begin{enumerate}[leftmargin=*]
    \item The environment generates an input--label pair $(x_t, y_t) \in \mathcal{X} \times \mathcal{Y}$ drawn from a distribution $P_t$.
    \item The learner observes $x_t$ and outputs a prediction $\hat{y}_t = f_{\theta_t}(x_t)$.
    \item The true label $y_t$ is revealed and the learner incurs loss $\ell(\hat{y}_t, y_t)$.
    \item The learner updates its parameters from $\theta_t$ to $\theta_{t+1}$.
\end{enumerate}
We focus on multi-class classification with cross-entropy loss, though the framework extends to regression.

\subsection{Drift model}

We adopt a drift-budget view in the spirit of variation-based analyses~\cite{HazanSeshadhri2007,BesbesEtAl2015}. Let $\mathrm{dist}(P,Q)$ denote a divergence between distributions on $\mathcal{X} \times \mathcal{Y}$, such as total variation distance. The cumulative drift is
\begin{equation}
\label{eq:drift-budget}
V_T = \sum_{t=2}^T \mathrm{dist}\big(P_t, P_{t-1}\big).
\end{equation}
This quantity captures both the frequency and magnitude of distributional changes.

It is also convenient to adopt a piecewise-stationary view with regimes $[t_{i-1}+1, t_i]$ and distributions $P^{(i)}$ that remain constant within each segment. Under suitable assumptions, such regime-based models and drift budgets are equivalent~\cite{BesbesEtAl2015}. Our theoretical discussion assumes that $V_T$ grows at most linearly with $T$ and is moderate relative to $\sqrt{T}$.

\subsection{Performance measures}

Empirically we evaluate \emph{prequential accuracy}: at each step $t$ we record $a_t = \mathbf{1}\{\hat{y}_t = y_t\}$ and consider both the final accuracy $a_T$ and the temporal average $T^{-1}\sum_{t=1}^T a_t$. This is standard in the data-stream literature~\cite{GamaEtAl2014}.

For theoretical purposes we use \emph{non-stationary regret}. Let $f_t^*$ denote a Bayes-optimal classifier under $P_t$. The regret of a strategy $\{f_{\theta_t}\}_{t=1}^T$ is
\[
R_T = \sum_{t=1}^T \ell\big(f_{\theta_t}(x_t), y_t\big)
      - \sum_{t=1}^T \ell\big(f_t^*(x_t), y_t\big).
\]
Regret compares the learner to an oracle that always uses the best classifier for the current distribution.

\section{Retrieval-Augmented Online Learning}
\label{sec:method}

We now describe the RAM-OL framework. We first outline the baseline online learner, then introduce the memory buffer and retrieval, followed by replay variants and their computational properties.

\subsection{Baseline: online MLP}

Our baseline is a one-hidden-layer multilayer perceptron (MLP) trained with online SGD. Given input $x_t \in \mathbb{R}^d$ and $C$ classes, the model computes
\[
 h_t = \sigma(W_1 x_t + b_1), \quad
 z_t = W_2 h_t + b_2, \quad
 p_t = \mathrm{softmax}(z_t),
\]
and minimises cross-entropy loss $\ell(p_t, y_t)$. The parameters $(W_1, b_1, W_2, b_2)$ are updated with a single SGD step per example. The hidden representation $h_t$ also serves as an embedding for similarity search in RAM-OL.

\subsection{Memory buffer and retrieval}

RAM-OL maintains a memory buffer $\mathcal{M}_t$ of fixed capacity $B$, initially empty. After processing each example $(x_t, y_t)$, the triple $(x_t, y_t, h_t)$ is inserted into $\mathcal{M}_t$. When the buffer is full, the oldest entry is evicted (FIFO). This keeps the buffer focused on recent history while potentially covering several regimes.

Before updating on $(x_t, y_t)$, RAM-OL retrieves up to $K$ neighbours:
\begin{enumerate}[leftmargin=*]
    \item Compute the embedding $h_t$.
    \item Optionally restrict candidates to indices $j$ with $t - j \leq H$ (time window).
    \item For eligible candidates, compute distances $d_j = \lVert h_t - h_j \rVert_2$.
    \item Select the $K$ nearest neighbours $N_t = \{(x_j, y_j, h_j)\}$.
\end{enumerate}
In our experiments we use exact search, which is sufficient for the moderate buffer sizes considered.

\subsection{Naïve replay}

The RAM-Naive variant augments the loss at time $t$ by averaging cross-entropy over the current example and its neighbours:
\[
L_t^{\text{naive}} =
\ell\big(f_{\theta_t}(x_t), y_t\big)
+
\frac{\beta}{|N_t|}
\sum_{(x_j, y_j, h_j) \in N_t}
\ell\big(f_{\theta_t}(x_j), y_j\big),
\]
with neighbour weight $\beta \geq 0$. When $\beta = 0$ or $N_t$ is empty, this reduces to the baseline. RAM-Naive treats all neighbours equally and does not discard stale or weakly similar memories.

\subsection{Gated replay}

The RAM-Gated variant adds three mechanisms aimed at robustness:
\begin{itemize}[leftmargin=*]
    \item \textbf{Time gate.} Only neighbours with $t - j \leq H$ are eligible, preventing very old examples from dominating updates when regimes have changed.
    \item \textbf{Similarity gate.} Using distances $d_j$, we compute similarities
    \(
    s_j = \exp(-d_j / \tau)
    \)
    and normalised weights
    \(
    w_j = s_j / \sum_{k \in N_t} s_k
    \).
    Neighbours with $w_j$ below a fraction $\rho$ of the maximum weight are discarded.
    \item \textbf{Gradient gate.} The current example receives weight $1$, while neighbours share total weight $\alpha \leq 1$. The gated loss is
    \[
    L_t^{\text{gated}} =
    \frac{1}{1+\alpha}
    \left(
    \ell\big(f_{\theta_t}(x_t),y_t\big)
    +
    \alpha \sum_{(x_j,y_j,h_j)\in N_t}
    w_j\,\ell\big(f_{\theta_t}(x_j),y_j\big)
    \right).
    \]
\end{itemize}
If no neighbours survive the gates, RAM-Gated defaults to the baseline update.

\subsection{Algorithm summary and complexity}

At each step $t$, RAM-OL:
\begin{enumerate}[leftmargin=*]
    \item receives $(x_t, y_t)$ and computes $h_t$;
    \item retrieves neighbours $N_t$ from $\mathcal{M}_{t-1}$;
    \item computes $L_t^{\text{naive}}$ or $L_t^{\text{gated}}$ and performs one SGD step;
    \item inserts $(x_t, y_t, h_t)$ into the buffer, evicting the oldest element if necessary.
\end{enumerate}
For buffer size $B$ and embedding dimension $d$, exact search costs $O(Bd)$ per step, comparable to the MLP forward pass. In practice, RAM-Gated increases total training time by roughly a factor of $1.5$--$2$ relative to the baseline (Section~\ref{sec:experiments}).

\section{Theoretical Considerations}
\label{sec:theory}

We briefly explain why retrieval-augmented updates can help under drift and outline their connection to regret bounds.

Consider a piecewise-stationary setting with regimes $[t_{i-1}+1, t_i]$ and distributions $P^{(i)}$. Within each regime the conditional risk of a predictor $f$ is
\[
\mathcal{L}^{(i)}(f)
  = \mathbb{E}_{(x,y)\sim P^{(i)}}\big[\ell(f(x),y)\big],
\]
and we denote by $f^{(i)*}$ a minimiser of $\mathcal{L}^{(i)}$. For convex, Lipschitz and strongly convex losses in a parameterised class, standard online convex optimisation guarantees $O(\sqrt{T})$ regret in the stationary case~\cite{Zinkevich2003}. In non-stationary settings, variation-based analyses show that suitably designed algorithms achieve regret of order $O(\sqrt{T} + V_T)$, where $V_T$ is the drift budget in~\eqref{eq:drift-budget}~\cite{HazanSeshadhri2007,BesbesEtAl2015}.

Retrieval becomes useful under an additional assumption of \emph{pattern recurrence}: when a regime $P^{(i)}$ appears (or reappears), the memory buffer contains examples whose distribution is close to $P^{(i)}$. Gradients computed on these neighbours approximate gradients under the new regime and can move the parameters towards $f^{(i)*}$ even before many fresh examples are observed. In effect, retrieval acts as a data-dependent warm start after change points.

Informally, if the learner can use memory to ensure that, after each change point, its parameters lie within a constant neighbourhood of $f^{(i)*}$, then the additional regret due to adaptation can be controlled by a term proportional to $V_T$ rather than by larger transient costs. The time and similarity gates help ensure that retrieved neighbours come from reasonably similar regimes, while the gradient gate prevents outdated examples from overwhelming the influence of new data.

Lower bounds in non-stationary optimisation show that no algorithm can do better than $\Omega(\sqrt{T} + V_T)$ regret in the worst case~\cite{BesbesEtAl2015}. In particular, if $V_T = \Theta(T)$, regret is necessarily linear. Thus RAM-OL does not alter the fundamental dependence on drift, but it can achieve favourable constants and faster adaptation in structured, recurring environments.

\section{Experimental Evaluation}
\label{sec:experiments}

We now evaluate RAM-OL on three real-world data streams. We address four questions:
\begin{enumerate}[leftmargin=*]
    \item Does retrieval-augmented memory improve accuracy and stability over a purely online baseline?
    \item How do naïve and gated replay variants compare across drift regimes?
    \item What is the computational overhead of RAM-OL?
    \item How do the gating components affect performance?
\end{enumerate}

\subsection{Datasets}

We use three publicly available binary classification streams that are standard in work on concept drift and data streams~\cite{WidmerKubat1996,GamaEtAl2014,Harries1999,BacheLichman2013}. Basic statistics are listed in Table~\ref{tab:datasets}.

\begin{table}[t]
    \centering
    \caption{Statistics of the three data streams.}
    \label{tab:datasets}
    \small
    \begin{tabular}{lrrrl}
        \toprule
        Dataset & \#Samples & \#Feat. & \#Cls. & Drift pattern \\
        \midrule
        ElecNormNew     & \num{45312}  &  8 & 2 & Strong, frequent \\
        ElectricityLoad & \num{100000} & 370 & 2 & Periodic + gradual \\
        AirlinesCSV     & \num{100000} &  8 & 2 & Weak, noisy \\
        \bottomrule
    \end{tabular}
\end{table}

\paragraph{ElecNormNew.}
An electricity price dataset derived from the New South Wales market~\cite{Harries1999,GamaEtAl2014}. Each example corresponds to a 30-minute interval with features describing time, day, demand and price indicators. The label indicates whether the price goes up or down relative to a moving average. The stream exhibits strong, frequent drift and seasonality.

\paragraph{ElectricityLoad.}
The Electricity Load Diagrams 2011--2014 dataset from the UCI repository~\cite{BacheLichman2013} records the consumption of 370 customers. We form a stream by using the instantaneous load vector as features and labelling whether the total load increases from time $t$ to $t+1$. The resulting stream is high dimensional and displays strong periodic patterns and slower long-term trends.

\paragraph{AirlinesCSV.}
A stream derived from airline on-time performance data provided by the U.S.\ Bureau of Transportation Statistics~\cite{BTSAirline}. Features include day of week, carrier and scheduled departure time; the label indicates whether the flight is delayed. Drift is relatively weak and noisy, driven by seasonal and weather factors and substantial label noise.

All streams are processed in temporal order without shuffling. Numerical features are standardised using running means and variances estimated from past data only.

\subsection{Evaluation protocol and baselines}

We use the prequential protocol~\cite{GamaEtAl2014}: at each step $t$, the model predicts on $x_t$, receives $y_t$, incurs loss, updates parameters and proceeds to $t+1$. There is no explicit train/test split.

We compare:
\begin{itemize}[leftmargin=*]
    \item \textbf{Baseline.} Online MLP without memory.
    \item \textbf{RAM-Naive.} RAM-OL with naïve replay (no time or similarity gating).
    \item \textbf{RAM-Gated.} RAM-OL with all three gates enabled.
\end{itemize}
The MLP architecture and learning rate schedule are shared across methods. RAM-OL hyperparameters (buffer size, number of neighbours, gating parameters) are fixed within each variant using a short validation prefix; we avoid dataset-specific tuning.

\begin{figure}[t]
    \centering
    \includegraphics[width=\textwidth]{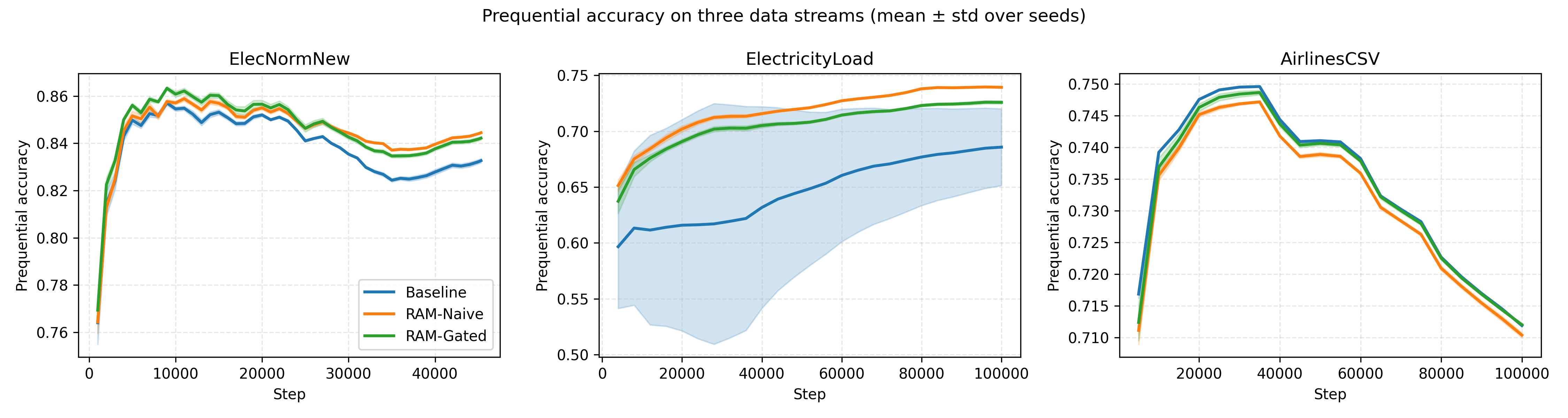}
    \caption{Prequential accuracy on the three data streams. Each subplot shows mean accuracy over three seeds with $\pm 1$ standard deviation bands for Baseline, RAM-Naive and RAM-Gated.}
    \label{fig:main}
\end{figure}

We run each configuration with three random seeds and report mean $\pm$ standard deviation of final and average prequential accuracy. Figure~\ref{fig:main} plots accuracy curves, and Table~\ref{tab:main} summarises the results.

\subsection{Main results}

\begin{table}[t]
    \centering
    \caption{Prequential accuracy of Baseline and RAM variants (mean $\pm$ standard deviation over three seeds).}
    \label{tab:main}
    \small
    \begin{tabular}{llcc}
        \toprule
        Dataset & Method & Final acc. & Avg.\ acc. \\
        \midrule
        AirlinesCSV & Baseline   & $0.7119 \pm 0.0002$ & $0.7338 \pm 0.0001$ \\
        AirlinesCSV & RAM-Gated  & $0.7120 \pm 0.0000$ & $0.7330 \pm 0.0006$ \\
        AirlinesCSV & RAM-Naive  & $0.7104 \pm 0.0005$ & $0.7315 \pm 0.0005$ \\
        \midrule
        ElecNormNew & Baseline   & $0.8327 \pm 0.0011$ & $0.8384 \pm 0.0012$ \\
        ElecNormNew & RAM-Gated  & $0.8422 \pm 0.0009$ & $0.8458 \pm 0.0013$ \\
        ElecNormNew & RAM-Naive  & $0.8445 \pm 0.0003$ & $0.8448 \pm 0.0009$ \\
        \midrule
        ElectricityLoad & Baseline   & $0.6857 \pm 0.0421$ & $0.6468 \pm 0.0823$ \\
        ElectricityLoad & RAM-Gated  & $0.7258 \pm 0.0018$ & $0.7052 \pm 0.0007$ \\
        ElectricityLoad & RAM-Naive  & $0.7393 \pm 0.0006$ & $0.7175 \pm 0.0009$ \\
        \bottomrule
    \end{tabular}
\end{table}

On \textbf{ElecNormNew}, both RAM variants consistently dominate the baseline across the stream (Figure~\ref{fig:main}, left). Gains in final and average accuracy are around one percentage point. The naïve variant is slightly stronger in terms of final accuracy, while the gated variant yields the highest average accuracy.

On \textbf{ElectricityLoad}, retrieval brings larger benefits. The baseline exhibits lower accuracy and substantial variability across seeds, whereas both RAM variants achieve markedly higher and more stable performance. RAM-Naive achieves the highest accuracy; RAM-Gated is slightly weaker but still clearly outperforms the baseline.

On \textbf{AirlinesCSV}, the three methods behave similarly. RAM-Gated closely tracks the baseline, while RAM-Naive is slightly worse on average. This stream exhibits weak signal and substantial noise; retrieval offers limited advantage and aggressive replay can overfit to misleading neighbours. The gated design prevents noticeable degradation.

Overall, RAM-OL provides clear improvements on strongly and periodically drifting streams, while RAM-Gated behaves as a conservative extension that does not harm performance on a noisy stream.

\subsection{Ablation study on ElecNormNew}

To understand the role of gating, we perform an ablation on ElecNormNew with the following variants: Baseline (no memory), RAM-Naive, Gated-full (all three gates), Gated-noTime (no time window), Gated-noSim (no similarity weighting) and Gated-noDecay (no gradient down-weighting).

\begin{figure}[t]
    \centering
    \includegraphics[width=0.75\textwidth]{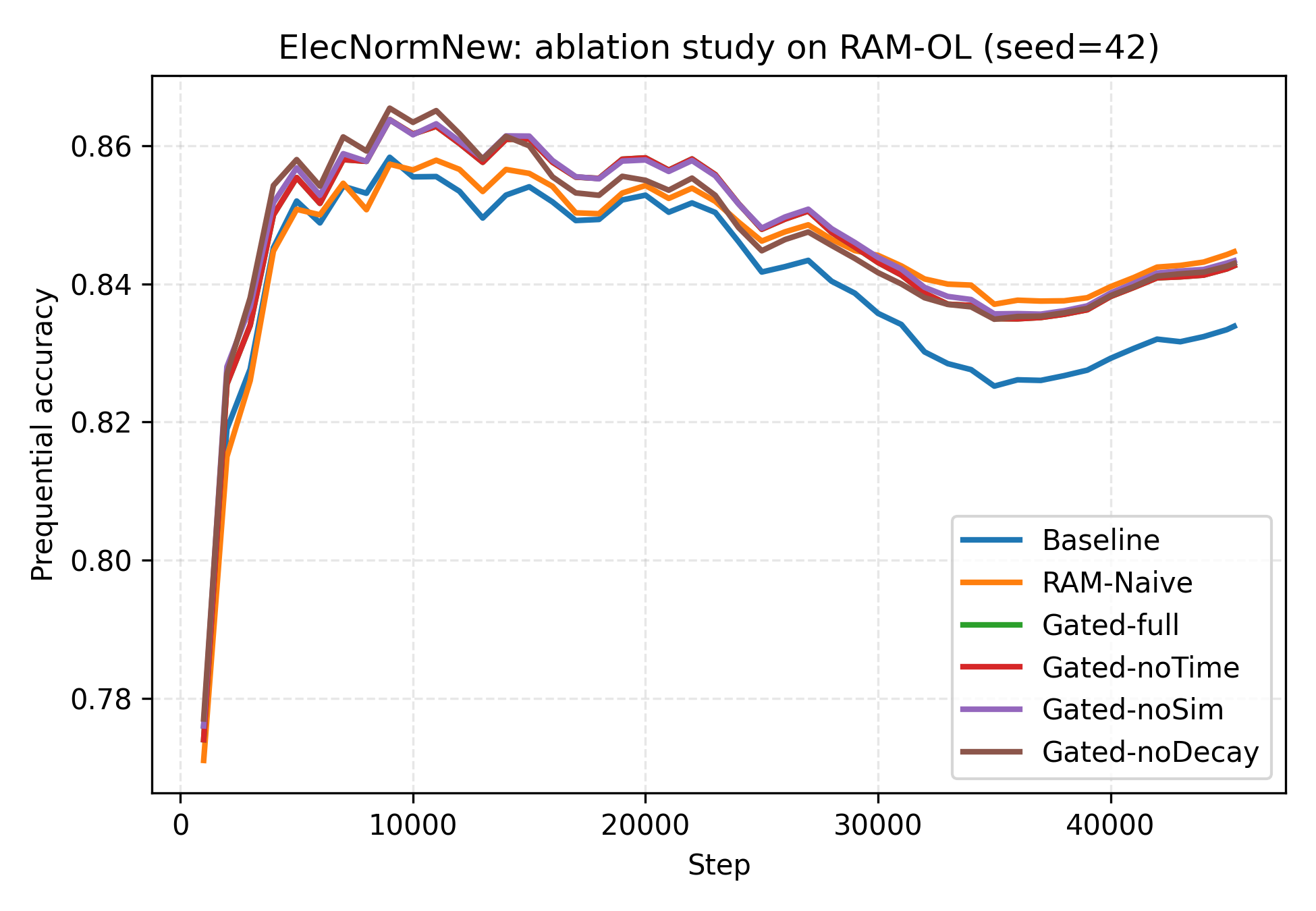}
    \caption{Ablation on ElecNormNew (seed 42). Prequential accuracy for Baseline, RAM-Naive and four gated variants.}
    \label{fig:ablation}
\end{figure}

Figure~\ref{fig:ablation} shows that all memory-based variants outperform the baseline. Differences among gated variants are small, indicating that RAM-OL does not rely on a fragile combination of hyperparameters; a broad family of gating choices yields similar benefits.

Table~\ref{tab:ablation} summarises final and average accuracy, neighbour coverage (fraction of steps with at least one retrieved neighbour) and neighbour label agreement (probability that a neighbour shares the current label).

\begin{table}[t]
    \centering
    \caption{Ablation on ElecNormNew: accuracy and memory statistics (single seed).}
    \label{tab:ablation}
    \small
    \begin{tabular}{lcccc}
        \toprule
        Method        & Final acc. & Avg.\ acc. & Coverage & Label match \\
        \midrule
        Baseline      & 0.8339 & 0.8397 & --   & --   \\
        RAM-Naive     & 0.8447 & 0.8446 & 1.00 & 0.76 \\
        Gated-full    & 0.8426 & 0.8466 & 1.00 & 0.75 \\
        Gated-noTime  & 0.8426 & 0.8466 & 1.00 & 0.75 \\
        Gated-noSim   & 0.8434 & 0.8472 & 1.00 & 0.75 \\
        Gated-noDecay & 0.8430 & 0.8464 & 1.00 & 0.75 \\
        \bottomrule
    \end{tabular}
\end{table}

Neighbour coverage is essentially one for all memory-based variants, and about three quarters of neighbours share the current label, confirming that the buffer provides locally consistent information. The consistent improvements over the baseline support the view that retrieval mainly refines gradients rather than radically changing learning dynamics.

\subsection{Efficiency and stability}

Retrieval introduces extra computation, but remains affordable in our setting. Table~\ref{tab:efficiency} reports runtime and seed-to-seed variability. The table is kept compact to avoid layout issues.

\begin{table}[t]
    \centering
    \caption{Efficiency and stability: runtime and variability over three seeds.}
    \label{tab:efficiency}
    \small
    \begin{tabular}{llccc}
        \toprule
        Dataset & Method & Final std & Avg.\ std & Time $\times$ \\
        \midrule
        AirlinesCSV & Baseline   & 0.0002 & 0.0001 & 1.00 \\
        AirlinesCSV & RAM-Gated  & 0.0000 & 0.0006 & 2.00 \\
        AirlinesCSV & RAM-Naive  & 0.0005 & 0.0005 & 1.43 \\
        \midrule
        ElecNormNew & Baseline   & 0.0011 & 0.0012 & 1.00 \\
        ElecNormNew & RAM-Gated  & 0.0009 & 0.0013 & 2.11 \\
        ElecNormNew & RAM-Naive  & 0.0003 & 0.0009 & 1.51 \\
        \midrule
        ElectricityLoad & Baseline   & 0.0421 & 0.0823 & 1.00 \\
        ElectricityLoad & RAM-Gated  & 0.0018 & 0.0007 & 2.29 \\
        ElectricityLoad & RAM-Naive  & 0.0006 & 0.0009 & 1.77 \\
        \bottomrule
    \end{tabular}
\end{table}

On ElecNormNew and ElectricityLoad, RAM-Naive is roughly $1.5$--$1.8\times$ slower and RAM-Gated about $2\times$ slower than the baseline. In return, memory substantially improves stability. For ElectricityLoad, the standard deviation of final accuracy drops from $0.0421$ (Baseline) to below $0.002$ for both RAM variants, a reduction of one to two orders of magnitude. Similar, though smaller, reductions appear on ElecNormNew. This variance reduction means fewer runs are needed for consistent performance and deployed models are less sensitive to initialisation.

\section{Discussion and Limitations}
\label{sec:discussion}

The results indicate that retrieval-augmented memory is most effective when drift is structured and recurring, such as the periodic patterns in electricity data and repeated market regimes seen in Elec-type streams~\cite{WidmerKubat1996,GamaEtAl2014,Harries1999}. In such settings, buffers often contain examples similar to the current regime and retrieval provides informative gradients. When drift is weak or dominated by noise, as in AirlinesCSV, retrieval yields only marginal gains; aggressive replay can even hurt, while the gated variant remains close to the baseline.

Naïve replay can sometimes achieve slightly higher accuracy, but it is less robust across datasets. RAM-Gated offers a safer default that improves or matches the baseline and markedly reduces variance. Buffer size and number of neighbours involve a trade-off between coverage and cost; in our experiments, moderate buffers and small $K$ suffice, but large-scale deployments would require approximate neighbour search and more elaborate memory management~\cite{BorgeaudEtAl2022}.

Our theoretical considerations rely on bounded drift and recurrence assumptions~\cite{GamaEtAl2014,BesbesEtAl2015}. In scenarios with abrupt, unprecedented changes, no method based solely on historical examples can avoid substantial transient error, and explicit model restarts or structural changes may be necessary. Finally, storing and replaying historical data raises privacy and robustness questions; designing memory mechanisms that respect privacy constraints and resist adversarial manipulation remains an important direction.

\section{Conclusion}
\label{sec:conclusion}

We introduced RAM-OL, a retrieval-augmented online learning framework for concept drift adaptation. By retrieving a small set of similar past examples and jointly updating on them and the current example, RAM-OL accelerates and stabilises adaptation to changing environments. A gated replay design balances the benefit of memory with protection against harmful interference from outdated data.

On three real-world data streams, RAM-OL consistently improves over a purely online MLP baseline on challenging electricity price and load streams and matches the baseline on a noisy airline delay stream, while greatly reducing variance across seeds. An ablation study shows that a broad family of gated variants outperforms the baseline, underscoring the robustness of the approach. Theoretical considerations link retrieval-augmented updates to regret bounds under bounded drift and highlight a threshold separating regimes where sublinear regret is possible from those where linear regret is unavoidable~\cite{Zinkevich2003,HazanSeshadhri2007,BesbesEtAl2015}.

Future work includes exploring larger backbones, extending RAM-OL to multi-task and reinforcement-learning settings, and studying privacy and security aspects of persistent memory in streaming learners.

\end{document}